\documentclass{article}

\usepackage{PRIMEarxiv}
\usepackage[labelfont=bf]{caption}
\usepackage{amsmath}
\usepackage[utf8]{inputenc} % allow utf-8 input
\usepackage[T1]{fontenc}    % use 8-bit T1 fonts
\usepackage{hyperref}       % hyperlinks
\usepackage{url}            % simple URL typesetting
\usepackage{booktabs}       % professional-quality tables
\usepackage{tabularx}

\usepackage{amsfonts}       % blackboard math symbols
\usepackage{nicefrac}       % compact symbols for 1/2, etc.
\usepackage{microtype}      % microtypography
\usepackage{lipsum}
\usepackage{fancyhdr}       % header
\usepackage{graphicx}       % graphics
\usepackage[affil-it]{authblk}
\usepackage{afterpage}   
\usepackage{multirow}
\allowdisplaybreaks % Allow page breaks in long captions
\usepackage{makecell}
\vspace{-15mm}

\newgeometry{textwidth=7in}

\title{An Auditable Agent Platform for Automated Molecular Optimisation}

\usepackage[affil-it]{authblk}


\author[1\thanks{Corresponding author email address: atabey@deltawave.fr}]{Atabey Ünlü}
\author[1]{Phil Rohr}
\author[1]{Ahmet Celebi}
\affil[1]{DeltaWave}

\begin{document}

\maketitle

% Lower the whitescape between abstract and authors

\begin{abstract}

Drug discovery frequently loses momentum when data, expertise, and tools are scattered, slowing design cycles. To shorten this loop we built a hierarchical, tool using agent framework that automates molecular optimisation. A Principal Researcher defines each objective, a Database agent retrieves target information, an AI Expert generates de novo scaffolds with a sequence to molecule deep learning model, a Medicinal Chemist edits them while invoking a docking tool, a Ranking agent scores the candidates, and a Scientific Critic polices the logic. Each tool call is summarised and stored causing the full reasoning path to remain inspectable. The agents communicate through concise provenance records that capture molecular lineage, to build auditable, molecule centered reasoning trajectories and reuse successful transformations via in context learning. Three cycle research loops were run against AKT1 protein using five large language models. After ranking the models by mean docking score, we ran 20 independent scale ups on the two top performers. We then compared the leading LLM’s binding affinity results across three configurations, LLM only, single agent, and multi agent. Our results reveal an architectural trade off, the multi agent setting excelled at focused binding optimization, improving average predicted binding affinity by 31\%. In contrast, single agent runs generated molecules with superior drug like properties at the cost of less potent binding scores. Unguided LLM runs finished fastest, yet their lack of transparent tool signals left the validity of their reasoning paths unverified. These results show that test time scaling, focused feedback loops and provenance convert general purpose LLMs into auditable systems for molecular design, and suggest that extending the toolset to ADMET and selectivity predictors could push research workflows further along the discovery pipeline.

\end{abstract}

\clearpage

% keywords can be removed
%\keywords{Drug design \and Transformer \and Generative adversarial networks \and Hepatocellular carcinoma}

\section{Introduction}

Molecular design is a fundamental and enduring effort across diverse scientific disciplines. Researchers rely on human expertise, experimental protocols, and physical and statistical models to design molecules for drug discovery, material development, and other applications \cite{du2024machine, pushkaran2024understanding,singh2024application, tang2024survey, yue2025machine}. However, the requirements for creating a viable product differ significantly across fields, making a universal, one-size-fits-all approach impractical. Consequently, tools and methodologies remain fragmented both across and within individual domains, as no single model can address the wide-ranging challenges inherent to molecular design \cite{maorchestrating, li2025drugpilot}. 

In drug discovery, these efforts are particularly evident. The Design-Make-Test-Analyze (DMTA) cycle typically spans multiple years and is labor- and resource-intensive. Despite enormous investments, most drug candidates fail in clinical trials, often owing to toxicity, poor specificity, or lack of efficacy, leading to multibillion-dollar losses \cite{chakraborty2025ai, zhang2025artificial,moran2005breakthrough}. Learning from these failures is crucial for ongoing drug development, yet the lengthy process and fragmentation of workflows can cause valuable information to be lost at various intersections in the pipeline \cite{hinostroza2025ai}. Drug design typically begins by identifying and characterizing a biological target (commonly a disease-related protein), which is then used to guide the development of compounds tailored to modulate the target’s activity \cite{scotti2022drug}. In cases where a target is not well-defined, ligand-based approaches leverage the pharmacokinetic properties of known active molecules, providing a pathway to discover new therapeutic agents \cite{tyagi2022pharmacophore}. The emergence of advanced artificial intelligence, particularly autonomous agentic systems, presents a promising new approach for navigating the design tools and processes \cite{goles2024peptide, vatansever2021artificial}. 

The fragmentation of data and siloed expertise in drug discovery are now being tackled by the introduction of Large Language Models (LLMs), which present a new approach for unifying varied information and workflows \cite{gangwal2024ai, al2024patient}. In contrast to specialized models, LLMs can be adapted for a multitude of tasks using in-context learning, which allows them to be fed task-specific examples during inference, thereby avoiding the need for extensive retraining \cite{bertsch2024context, jeon2024information, edwards2023synergpt, dong2022survey}. This enables a single, robust model to seamlessly transition between different contexts, such as interpreting biological data and proposing chemical alterations \cite{qi2025improving, schimunek2025mhnfs, swanson2024virtual, lu2024ai}. Additionally, the principle of test-time-compute allows these models to undertake more complex reasoning by repeatedly generating, assessing, and improving potential solutions \cite{gottweis2025towards,novikov2025alphaevolve, zhang2025survey}. This ability for immediate adaptation and more profound, iterative reasoning offers a potent set of tools to clear the hurdles in the conventional DMTA cycle \cite{abunasser2024large, ghiandoni2024augmenting}. Leveraging these characteristics facilitates the creation of more cohesive and automated systems for molecular design that can maneuver through the intricate, multi-objective terrain of drug discovery \cite{m2024augmenting,narayanan2024aviary}. 

In this study, we present and evaluate a multi-agent system designed to systematically investigate the performance of LLMs in a tool-augmented, target-based molecular design task. Our primary goal was to construct a workflow that emulates the sequential, multi-disciplinary process of a computational drug discovery campaign, where complex problems are decomposed into specialized sub-tasks. Our system operates through iterative Research Cycles (Figure 1), initiated by a Principal Researcher agent that defines the overarching objective. Control then proceeds sequentially: a Database Agent first retrieves foundational target data, which is used by an AI Expert to generate de novo molecular scaffolds. These candidates are subsequently passed to a Medicinal Chemist agent, which leverages a suite of computational tools, including molecular docking and property calculators, to iteratively optimize the structures. Finally, a Ranking Agent synthesizes all results to prioritize the most promising compounds before the cycle concludes. By formalizing this structured collaboration, we dissect how different agentic architectures leverage fragmented tools to navigate the multi-objective landscape of molecular optimization. Our findings confirm that LLMs, when equipped with computational tools, can perform targeted molecular optimization. We reveal that the choice of agentic architecture critically influences the strategic outcome: the multi-agent configuration is highly effective at maximizing a single objective like predicted binding affinity, while a single-agent architecture naturally balances potency with broader drug-like properties, highlighting different paths toward viable candidate discovery.

\subsection{Related Work}

\paragraph{Tool-Augmented Scientific Agents.} The capabilities of Large Language Models (LLMs) in scientific discovery are significantly amplified when they are augmented with external tools. This approach shifts the paradigm from creating a single, monolithic model to developing a proficient "tool-user" that can leverage specialized functions to gather information, perform complex calculations, and interact with other systems \cite{schick2023toolformer, basu2024api, qin2023toolllm}. By integrating with external databases, APIs, and computational models, these tool-augmented agents can overcome the inherent limitations of pre-trained models, such as knowledge cutoffs and the propensity for generating inaccurate information \cite{das2024mathsensei, paranjape2023art}. In fields like chemistry and drug discovery, this means an agent can access chemical databases to retrieve compound information, execute computational models for tasks like molecular docking, and analyze biological data \cite{mcnaughton2024cactus, ma2024sciagent, gao2025pharmagents, ghafarollahi2024protagents}. This augmentation not only enhances the accuracy and reliability of the agent's outputs but also enables a more dynamic and interactive research process, accelerating the pace of scientific breakthroughs by bridging the gap between computational analysis and domain-specific challenges \cite{schmidgall2025agent}. 

\paragraph{Multi-Agent Systems.} To tackle the multifaceted challenges inherent in scientific research, the focus is shifting from single-agent systems to multi-agent systems (MAS) \cite{ghafarollahi2024protagents, li2024survey, han2024llm, du2024survey}. MAS shows better performance in complex, dynamic environments by distributing tasks among specialized agents that collaborate to achieve a common goal. This approach offers several advantages, including enhanced scalability, adaptability, and resilience, as the workload is distributed and the failure of a single agent does not compromise the entire system\cite{acha2025cooperative, xie2017multi, wu2011online}. This collaborative intelligence allows for the integration of diverse expertise and data sources, mirroring the cross-disciplinary nature of scientific research and accelerating breakthroughs by enabling a more comprehensive and efficient exploration of the problem space \cite{sun2025multi, krishnan2025advancing,li2024agent, tran2025multi}. Recent frameworks like AutoGen and CrewAI are being developed to facilitate the creation and orchestration of these collaborative AI teams \cite{wu2024autogen, taulli2025crewai}. 

\paragraph{Architectures of Agent Collaboration.} The effectiveness of a multi-agent system is heavily dependent on its collaborative architecture, which defines how agents interact and coordinate their actions \cite{maldonado2024multi}. Common models include centralized , hierarchical \cite{wang2024hierarchical}, and decentralized structures \cite{yang2025agentnet}. In a centralized model, a single "supervisor" or "orchestrator" agent manages and delegates tasks to all other agents, which is effective for tightly controlled workflows but can create bottlenecks \cite{maldonado2024multi}. Hierarchical architectures, in contrast, create layers of control, with top-level agents delegating to intermediate managers, enhancing scalability and allowing for more localized decision-making \cite{liu2025pc}. The choice of architecture depends on the specific problem, with some systems employing hybrid models that combine elements of different structures to optimize for both control and adaptability. Frameworks are emerging to help developers design and implement these varied collaborative workflows, enabling the creation of sophisticated, self-organizing agent networks \cite{liu2024dynamic, qian2024scaling, tran2025multi}.

\begin{figure*}

    \centering
    \includegraphics[width=\textwidth]{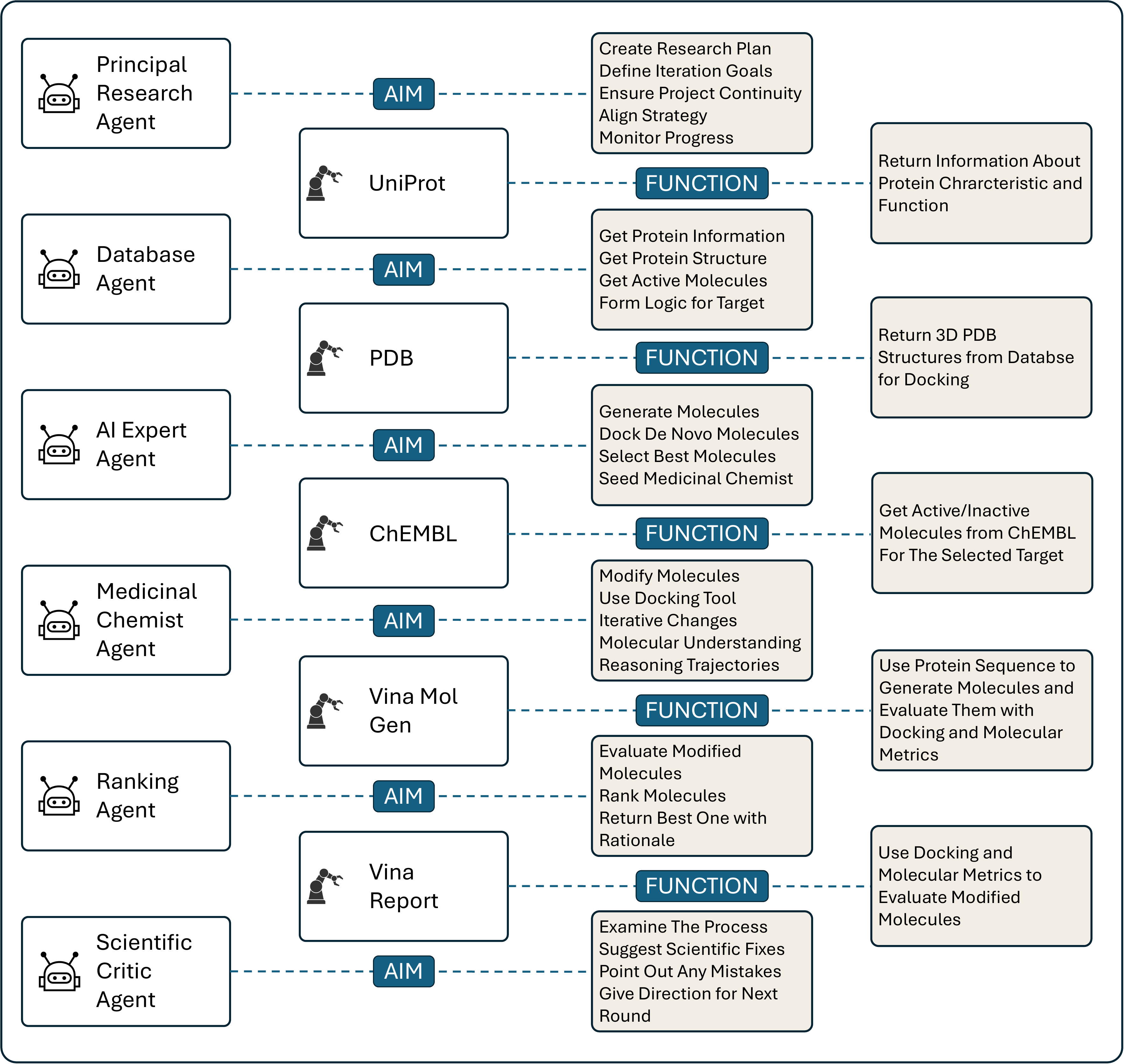}

    \caption{This figure depicts the multi-agent architecture, which employs six agents and five tools. Three tools, UniProt \cite{uniprot2025uniprot}, PDB \cite{burley2025updated} and ChEMBL \cite{zdrazil2025fifteen}, are accessed through single API calls, whereas the other two, Vina-Mol-Gen and Vina-Report, are composite workflows that bundle multiple tools into a single invocation.
}
    \label{fig:main}
\end{figure*}

\section{Method}

\subsection{Multi Agent System}

Our framework is implemented as a hierarchical multi-agent system operating under sequential conversation model. It is designed to automate the molecular optimization phase of de novo drug discovery. The architecture is orchestrated by a Principal Researcher agent, which defines high-level objectives and coordinates the tasks of specialized subordinate agents in a sequential workflow. Each cycle is initiated by the Principal Researcher, which then passes control and context to the appropriate agent in a predefined order. A core design principle is the integration of LLM-driven reasoning with deterministic, rule-based computational validation. This hybrid approach ensures that while agents can leverage the flexible reasoning of LLMs, every proposed candidate is systematically evaluated against a consistent set of scientific criteria, enforcing rigor throughout the workflow. Methodological implementation and experiment details can be found in Supplementary Information S1-4.

\subsection{Agents}

\paragraph{Principal Researcher Agent} The Principal Researcher acts as the orchestrator of the multi-agent system. Its primary role is to define the overarching research strategy, set specific objectives for each iteration, and coordinate the activities of all specialist agents. It does not use tools directly but instead synthesizes the data and findings from other agents to produce a comprehensive summary at the end of each cycle, providing clear, actionable directives to guide the subsequent iteration and ensure project continuity. 

\paragraph{Database Agent} This agent is responsible for foundational data retrieval. Equipped with tools to query biochemical databases like UniProt \cite{uniprot2025uniprot}, PDB\cite{burley2025updated}, and ChEMBL \cite{zdrazil2025fifteen}, it gathers essential information such as the target protein's structure, its sequence, and data on known inhibitors and their activities. It provides this structured data to the team, establishing the factual basis for the subsequent design and optimization work.

\paragraph{AI Expert Agent} This agent serves as the de novo design engine of the system. Its sole function is to generate novel molecular starting points using a generative AI tool. It takes the project's target information and proposes initial, structurally plausible candidate molecules. These molecules then serve as the initial chemical matter for the optimization-focused agents to refine.

\paragraph{Medicinal Chemist Agent} The agent is a domain expert agent tasked with the iterative, goal-directed optimization of molecules. It receives candidate structures from the AI Expert and systematically proposes modifications to improve target binding. Crucially, it uses a docking tool to evaluate the potency of its own modifications, creating a tight design-test-learn loop. This process continues until a satisfactory improvement in docking score is achieved.

\paragraph{Ranking Agent} This agent functions as a data synthesizer and prioritizer. It aggregates all generated molecules and their associated data, including docking scores, chemical properties, and modification histories, from the other agents. As an LLM-based evaluator, it ranks candidates by quantitative multi-parameter objectives while incorporating qualitative intuition from its pretraining on vast world knowledge, yielding a final, ranked list of the most promising compounds for the Principal Researcher.

\paragraph{Scientific Critic} The Scientific Critic acts as a quality control mechanism within the system. It is a LLM-as-Judge agent that does not participate in generation or analysis but instead scrutinizes the reasoning and outputs of the other agents. Its purpose is to identify potential logical inconsistencies, flawed scientific assumptions, or deviations from the project's objectives, ensuring the overall rigor and validity of the research process.

\subsection{Workflows and Tools}

Our agent framework is built upon a modular library of computational tools and integrated workflows, which are summarized in Table 1. The system is designed with a hierarchical structure: individual tools perform discrete, well-defined tasks, such as retrieving data from a specific database, executing a docking simulation, or calculating a single molecular property. These tools are powered by established scientific software, including RDKit \cite{Landrum_2013} for cheminformatics calculations, Prot2Mol \cite{unlu2024prot2mol} for molecule generation, AutoDock Vina \cite{trott2010autodock} for docking, and PLIP \cite{salentin2015plip} for interaction analysis. To automate more complex, multi-step operations, these atomic tools are chained together into cohesive workflows. For instance, the $vina\_mol\_gen$ workflow orchestrates a full de novo design cycle, it first generates novel structures, filters them based on drug-like properties, docks the most promising candidates, and finally analyses their binding modes. This architecture allows agents to operate at different levels of abstraction, from executing a single calculation to launching a comprehensive design-and-evaluate campaign with a single command.

\renewcommand{\theadfont}{\fontsize{7}{11}}
\begin{table}
\setlength\tabcolsep{2.4pt}
    \caption{Computational workflows and auxiliary tools implemented in our structure-based drug‐design pipeline.}
    \centering
    \fontsize{7}{11}\selectfont
    \makebox[\linewidth]{
    \begin{tabular}{llll}
    \hline
    \textbf{Name} & \textbf{Description} & \textbf{Underlying Method / Library} & \textbf{Primary Function} \\
    \hline
    \textbf{Workflows} \\
    \hline
$vina\_mol\_gen$ &
\thead[l]{An end-to-end pipeline that \\ generates de novo molecules from a \\ protein sequence, filters them, and \\ evaluates predicted binding affinity.} &
\thead[l]{Prot2Mol \cite{unlu2024prot2mol}, RDKit \cite{Landrum_2013}, \\ AutoDock \\ Vina \cite{trott2010autodock}, PLIP \cite{salentin2015plip}} &
\thead[l]{De novo molecule generation \\ and initial screening.} \\
$vina\_report$ &
\thead[l]{Evaluates a user-provided list of \\ molecules by performing docking \\ and analyzing protein-ligand \\ interactions.}&
\thead[l]{AutoDock Vina \cite{trott2010autodock}, \\ PLIP \cite{salentin2015plip}, RDKit\cite{Landrum_2013}}&
\thead[l]{Evaluate and score existing \\ molecules.}\\
$search\_chembl\_activity$ &
\thead[l]{Retrieves known active and inactive \\ compounds for a given protein target \\ from the ChEMBL database and \\ analyzes the binding profiles}&
\thead[l]{ChEMBL API \cite{zdrazil2025fifteen}, \\ Autodock Vina \cite{trott2010autodock}, \\ PLIP \cite{salentin2015plip}}&
\thead[l]{Gather initial SAR data and \\ reference compounds.}\\
    \hline
    \textbf{Tools} \\
    \hline
$search\_uniprot$ &
\thead[l]{Fetches protein information \\ (sequence, function, identifiers) from \\ the UniProt database.} &
UniProt API \cite{uniprot2025uniprot}&
\thead[l]{Retrieve target protein \\ metadata.}\\
$get\_pdb\_file$&
\thead[l]{Downloads a specified protein \\ structure file from the Protein Data \\ Bank.}&
PDB API \cite{burley2025updated}&
\thead[l]{Obtain the 3D structure of the \\ target protein.}\\
$afdb\_call$&
\thead[l]{Fallback call to AlphaFoldDB if no \\ pdb file found in Protein Data Bank}&
AF2DB API \cite{varadi2024alphafold}&
\thead[l]{Obtain AF2 prediction for \\ target protein.}\\
$generate\_molecules$&
\thead[l]{Generates novel small-molecule \\ SMILES based on a target protein's \\ amino acid sequence.}&
Prot2Mol \cite{unlu2024prot2mol}&
\thead[l]{Core generative engine \\ for target based de novo molecule \\ design.}\\
$run\_vina\_docker$&
\thead[l]{Performs molecular docking of one \\ or more ligands to a protein target, \\  returning binding energy scores.}&
AutoDock Vina \cite{trott2010autodock}&
\thead[l]{Predict predicted binding affinity and \\ pose.}\\
$get\_plip\_report$&
\thead[l]{Analyzes and reports the \\ non-covalent interactions between a \\ protein and a docked ligand.}&
PLIP \cite{salentin2015plip}&
\thead[l]{Characterize the binding \\ mode.}\\
validity, QED \cite{bickerton2012quantifying}, SA \cite{ertl2009estimation}, logP&
\thead[l]{Calculate fundamental molecular \\ properties: SMILES validity, \\ Quantitative Estimate of \\ Drug-likeness, and Synthetic \\ Accessibility.}&
RDKit \cite{Landrum_2013}&
\thead[l]{Filter and assess molecule \\ quality}. \\
\hline
    
    \label{tab:table1}
    \end{tabular}}

\end{table}

\subsection{Agent Schedule}

The operational workflow of our system is structured into discrete, sequential \textit{Research Cycles}, each initiated and supervised by the \textit{Principal Researcher Agent}. This cycle methodically emulates the logical stages of a scientific investigation, and ensures a structured progression from data gathering to candidate selection. Following the initial task definition by the \textit{Principal Researcher}, the \textit{Database Agent} executes foundational data retrieval, providing the necessary protein and ligand context for the subsequent steps. This information is then passed to the \textit{AI Expert}, which generates a set of de novo molecular scaffolds. These initial candidates are then handed off to the \textit{Medicinal Chemist Agent} for iterative, goal-directed optimization, where each proposed modification is immediately tested using integrated docking tools. The resulting pool of optimized molecules is systematically evaluated and prioritized by the \textit{Ranking Agent}, and the \textit{Scientific Critic} provides a final layer of examination over the process and its outputs. The cycle concludes when the curated results are returned to the Principal Researcher, which then assesses the outcomes to determine whether the objectives have been met or if a new cycle with refined objectives is required. This iterative structure allows the system to build upon previous findings and progressively converge on high-quality solutions.

\subsection{Research Cycles}

Each experiment proceeds through three sequential research cycles. Every cycle is an turn-based conversation in which only the agents listed for that stage are allowed to act; all others remain silent. The Principal Researcher (PR) always opens the cycle by setting objectives and closes it by summarising outcomes and issuing marching orders for the next round. Summarizing conversation history within in-context learning environments has been shown to boost exploration and, in turn, sharpen subsequent exploitation \cite{krishnamurthy2024can}. The composition and purpose of each cycle are shared in Table 2.

At the end of each cycle the PR publishes a concise summary plus a structured provenance record. Only this distilled context is passed forward, preventing context-window bloat while preserving all essential lineage information. At the close of each cycle the Principal Researcher issues a brief narrative summary accompanied by a structured provenance linkage. This distilled package, rather than the full raw conversation, becomes the sole context handed to the next round, preventing context-window bloat while preserving every lineage link. Tool use is similarly regimented. The Database agent anchors the project only once, in Cycle 1, by querying UniProt \cite{uniprot2025uniprot}, PDB \cite{burley2025updated}, and ChEMBL \cite{zdrazil2025fifteen} for target sequences, structures, and known ligands. During that same first pass the AI Expert is the only component allowed to invoke Prot2Mol \cite{unlu2024prot2mol}, seeding the campaign with de-novo scaffolds; it is then not used for the remainder of the run. From that point onward the Medicinal Chemist drives optimization in every cycle, docking each analogue with AutoDock Vina \cite{trott2010autodock} and calculating core RDKit \cite{Landrum_2013} properties. A dedicated Ranking agent folds these scores into a multi-parameter desirability index and returns the current top candidates, while the Scientific Critic reviews the chain of reasoning and tool outputs, contributing corrective commentary but no molecules of its own. The loop terminates automatically after three passes. This phased, shrinking-team design front-loads broad exploration and then concentrates resources in depth, enabling the system to generate, refine, and finalise candidate chemotypes in a transparent and auditable fashion.

\renewcommand{\theadfont}{\fontsize{7}{11}}
\begin{table} [h]
\setlength\tabcolsep{2.8pt}
    \caption{Agent composition and objectives across the three research cycles.}
    \centering
    \fontsize{7}{11}\selectfont
    \makebox[\linewidth]{
    \begin{tabular}{lll}
    \hline
    \textbf{Cycle} & \textbf{Active agents (execution order)} & \textbf{Primary purpose} \\
    \hline
    \thead[l]{Cycle 1 \\ (Seeding)} & 
    \thead[l]{PR $\rightarrow$ Database $\rightarrow$ AI Expert \\ $\rightarrow$ Medicinal Chemist $\rightarrow$ Ranking \\ $\rightarrow$ Scientific Critic} &
    \thead[l]{Assemble target context, generate \textit{de novo} scaffolds via the \\ sequence-to-molecule model, perform an initial round of \\ structure–activity optimisation, prioritise candidates, and audit \\ the reasoning trajectory.} \\[2pt]

    \thead[l]{Cycle 2 \\ (Optimising)} & 
    \thead[l]{PR $\rightarrow$ Medicinal Chemist \\ $\rightarrow$ Ranking $\rightarrow$ Scientific Critic} &
    \thead[l]{Iteratively refine the best candidates from Cycle 1; no new \\ scaffolds are introduced. Docking feedback, property filters, and \\ critic comments steer focused exploration around existing \\ chemotypes.} \\[2pt]

    \thead[l]{Final Cycle \\ (Selection)} & 
    \thead[l]{PR $\rightarrow$ Medicinal Chemist $\rightarrow$ Ranking} &
    \thead[l]{Carry out a final polishing pass, rank all surviving molecules, \\ and compile the closing report. The critic is omitted to \\ streamline closure; any residual issues are flagged by the PR.} \\
    \hline
    \label{tab:research_cycles}
    \end{tabular}}
\end{table}

\section{Results}

Here, we report results from a multi-agent pipeline designed to optimize molecules using signals from various computational tools. Detailed descriptions of the experimental procedures and agent properties are provided through Supplementary Information 1-5. The pipeline consisted of agents that leveraged a total of five distinct tools (except in LLM-only control experiments), with each experimental condition replicated three times to ensure robustness and reproducibility. Initially, five distinct large language models (LLMs) were evaluated, after which the two top-performing models were selected for further scaling of molecule generation tasks. This approach was employed to enhance experimental efficiency and reduce computational costs. Following scale-up, the best-performing LLM within the multi-agent setup was identified based on two key metrics: the average docking score and the average absolute change in docking scores for the generated molecules. The prompts of the system were tuned on Sonnet-3.7 and then applied unchanged to the other LLMs, which may lead to comparatively lower performance for models whose optimal prompting differs \cite{chang2025transfer, miehling2024evaluating, gozzi2024comparative}. All experiments were executed using a single Nvidia A100 40GB GPU, with parallelization strategies employed to minimize GPU resource usage and overall computational cost. 

We comprehensively evaluated our multi-agent system across various dimensions, including molecular generation capability, molecular reasoning proficiency, and efficacy in utilizing computational tools. This study serves as a feasibility assessment to critically evaluate how different agentic architectures navigate the multi-objective challenge of drug design. We aim to highlight not only the potential for automation but also to characterize the distinct strategic biases and trade-offs inherent in each approach. To the best of our knowledge, this study is the first open-source investigation to dissect and compare the capabilities of deploying LLMs in baseline, single-agent, and multi-agent configurations on target based molecular optimization tasks, providing a crucial analysis of their respective strengths and weaknesses.

\subsection{Overall Comparison of Different LLMs}

To compare the multi-agent capabilities of five closed-source LLMs (from OpenAI, Anthropic, and Google), we ran each model through the same three-run experiment and judged them on three axes: (i) the chemical quality of the molecules they produced, (ii) standard drug-likeness metrics, and (iii) target-specific docking performance.

We first generate candidate ligands with Prot2Mol \cite{unlu2024prot2mol}, a deep-learning model that turns the target’s amino-acid sequence into de novo molecules. A downstream Medicinal Chemist agent then edits these molecules, based on its internal knowledge and learned from in-context information, in successive cycles making tool calls to Autodock Vina \cite{trott2010autodock}, aiming to improve their docking scores. All results discussed below therefore come from molecules that were created by Prot2Mol and subsequently refined by the agent. In this study, we selected AKT1 as the target protein to evaluate the different LLMs, primarily due to its well-characterized structural and functional properties \cite{mroweh2021targeting,manning2007aktpkb,matheny2009aktivation,martorana2021akt,george2022akt1,stahl2006integrating,chen2018differential,thomas2002phdomain,turner2023capivasertib,addie2013azd5363,lin2012atpswitch}. While its performance in optimizing docking scores made \textit{claude-3-7-sonnet-20250219} the clear choice for our subsequent architectural comparisons, it is crucial to note that this single metric does not capture the full picture of a molecule's therapeutic potential. 

In Table 3, we compare the MAS performance across different LLMs by examining the total number of molecules generated over three replicate runs. Each model was prompted to propose 10 optimized molecules at the end of each run; however, the table evaluates all molecules generated by the LLMs throughout the replicates. Among the tested models, \textit{claude-sonnet-4-20250514} generated the most molecules, while GPT-4.1 generated the fewest. Importantly, all models produced valid and unique molecular SMILES representations. Upon comparing the generated molecules to known active AKT1 molecules, all generated structures were confirmed as novel. On average, \textit{claude-3-7-sonnet-20250219} produced molecules most structurally similar to known AKT1 inhibitors, achieving an average maximum pairwise Tanimoto similarity of 0.458. In addition to structural similarity, we also employed the Frechet ChemNet Distance (FCD) score to evaluate (lower is better) the physicochemical similarity of the generated molecules to known AKT1 inhibitors. This analysis revealed that the \textit{claude-sonnet-4-20250514} model produced molecules with the highest physicochemical similarity, narrowly surpassing \textit{claude-3-7-sonnet-20250219}. Regarding drug-likeness metrics, \textit{openai/o3-2025-04-16} performed best in terms of average QED scores and Lipinski rule compliance, while \textit{gemini/gemini-2.5-pro-preview-06-05} \cite{team2023gemini} achieved the highest performance in SA scores and Lipinski passes. 

\renewcommand{\theadfont}{\fontsize{7}{11}}
\begin{table}[h]
    \setlength\tabcolsep{4.2pt}   % tight columns
    \caption{Performance comparison of the different LLMs based on MAS runs evaluated on generation quality, drug-likeness \& properties and target specific performance.}
    \centering
    \fontsize{7}{11}\selectfont
    \makebox[\linewidth]{%
    \begin{tabular}{lccccc}
    \hline
    & \textbf{Sonnet-3.7} & \textbf{Sonnet-4} & \textbf{o3} & \textbf{Gemini} & \textbf{gpt-4.1} \\
    \hline
    \textbf{Generation Quality} \\ \hline
    Molecules Attempted (N)\textsuperscript{*} & 75 & 101 & 44 & 43 & 29 \\
    Validity (\%)                              & 1.000 & 1.000 & 1.000 & 1.000 & 1.000 \\
    Uniqueness (\% of Valid)                   & 1.000 & 1.000 & 1.000 & 1.000 & 1.000 \\
    Novelty (\% vs.\ Actives)                  & 1.000 & 1.000 & 1.000 & 1.000 & 1.000 \\
    Internal Diversity                         & 0.764\,$\pm$\,0.056 & 0.779\,$\pm$\,0.043 & \textbf{0.800}\,$\pm$\,0.031 & 0.800\,$\pm$\,0.026 & 0.714\,$\pm$\,0.090 \\
    Similarity to known inhibitors             & \textbf{0.458}\,$\pm$\,0.163 & 0.376\,$\pm$\,0.180 & 0.289\,$\pm$\,0.076 & 0.386\,$\pm$\,0.220 & 0.259\,$\pm$\,0.027 \\
    FCD                                         & 26.148 & \textbf{25.152} & 34.451 & 32.181 & 38.774 \\
    \hline
    \textbf{Drug-Likeness \& Properties} \\ \hline
    QED (Mean\,$\pm$\,SD)                      & 0.395\,$\pm$\,0.177 & 0.606\,$\pm$\,0.167 & \textbf{0.646}\,$\pm$\,0.105 & 0.638\,$\pm$\,0.173 & 0.592\,$\pm$\,0.108 \\
    Lipinski Ro5 Pass (\%)                     & 56.00 & 92.00 & \textbf{100} & \textbf{100} & 96.50 \\
    SA Score (Mean\,$\pm$\,SD)                 & 3.264\,$\pm$\,0.403 & 3.234\,$\pm$\,0.606 & 2.987\,$\pm$\,0.687 & \textbf{2.772}\,$\pm$\,0.450 & 3.200\,$\pm$\,0.673 \\
    LogP (Mean\,$\pm$\,SD)                     & 3.667\,$\pm$\,1.599 & 2.600\,$\pm$\,1.328 & 2.782\,$\pm$\,1.135 & 3.045\,$\pm$\,0.972 & 3.312\,$\pm$\,0.938 \\
    \hline
    \textbf{Target-Specific Performance} \\ \hline
    Docking Score (Mean\,$\pm$\,SD)            & $-10.973$\,$\pm$\,1.573 & $-9.364$\,$\pm$\,1.209 & $-9.811$\,$\pm$\,1.283 & $-8.304$\,$\pm$\,1.538 & $-9.222$\,$\pm$\,0.839 \\
    Hits (Dock $< -9.0$ kcal/mol) (\%)         & 96.00 & 60.75 & 84.09 & 37.21 & 62.07 \\
    \hline
    \end{tabular}}
{\raggedright * We did not enforce a fixed exploration depth; each LLM was free to generate and discard as many intermediate structures as it deemed necessary before converging on its ten final proposals. Therefore, “Molecules Attempted (N)” reports the total number of candidate SMILES evaluated across all optimization steps (and all three replicates) for each model, and the values naturally differ from one LLM to the next.
\par} 
\vspace{0.5em}
\hrule
    \label{tab:llm_mas_metrics}
\end{table}

The average LogP value for known active AKT1 molecules was 4.006, with \textit{claude-3-7-sonnet-20250219} generating molecules closest to this value, indicating better alignment with known inhibitor profiles. For target-specific performance, we compared the models based on the average docking score and the proportion of hits (defined as molecules achieving docking scores below -9.0 kcal mol\textsuperscript{-1}). When evaluated primarily on predicted binding affinity, \textit{claude-3-7-sonnet-20250219} emerged as the top performer, achieving an average docking score of -10.973 kcal mol\textsuperscript{-1} and a 96\% hit ratio. Other models, such as openai/o3-2025-04-16, produced molecules with better drug-likeness profiles (e.g., 100\% Lipinski compliance), illustrating a trade-off between predicted potency and other desirable properties even at the model-selection stage. While this made it the ideal candidate for our architectural comparison, we acknowledge that a singular focus on predicted binding affinity can obscure other vital drug properties. Given our initial focus on maximizing predicted binding affinity as a primary test of optimization capacity, \textit{claude-3-7-sonnet-20250219} was selected as the most effective model for this specific task to carry forward into our architectural comparisons, and our subsequent analysis will explore how this potent engine behaves under different architectural constraints.

Figure 2 shows how each LLM-driven agent team reshaped its initial chemotypes over three optimisation rounds. Across all models except \textit{gemini/gemini-2.5-pro-preview-06-05}, the mean docking scores fell below the -9 kcal mol\textsuperscript{-1} threshold, confirming that the MAS pipeline consistently steers molecules toward tighter AKT1 binding. The improvement is most striking for \textit{claude-3-7-sonnet-20250219}, whose average docking score drops by roughly 3 kcal · mol\textsuperscript{-1} about twice the gain achieved by \textit{gpt-4.1-2025-04-14} and clearly above the other systems corresponding to a ~30 \% boost relative to its de-novo baseline. Crucially, this potency jump is obtained with only modest trade-offs: QED declines by just 5–11 \% for every model, and two models (\textit{openai/o3-2025-04-16} and \textit{gemini/gemini-2.5-pro-preview-06-05}) even reduce synthetic-accessibility scores, meaning their optimised compounds should be easier to synthesize. Panel D shows that the LLMs explored very different numbers of intermediates thorugh iterations to their ten final molecules (e.g., \textit{claude-sonnet-4-20250514} evaluates 30 candidates while \textit{gpt-4.1-2025-04-14} tests only 12), yet exploration breadth does not track with performance, underscoring that search quality drives success. Finally, model-specific biases emerge: \textit{gpt-4.1-2025-04-14} protects QED at the expense of SAS, whereas \textit{gemini/gemini-2.5-pro-preview-06-05} performs with a slightly smaller docking gain but better synthesizability. Taken together, the figure demonstrates that the multi-agent framework can rapidly improve predicted binding affinity. However, it also reveals the challenge of balancing this optimization with other properties. The framework highlights \textit{claude-3-7-sonnet-20250219} as the most efficient model for translating the MAS's tool-driven signals into dramatic gains in predicted binding affinity.

\begin{figure*}[h]

    \centering
    \includegraphics[width=\textwidth]{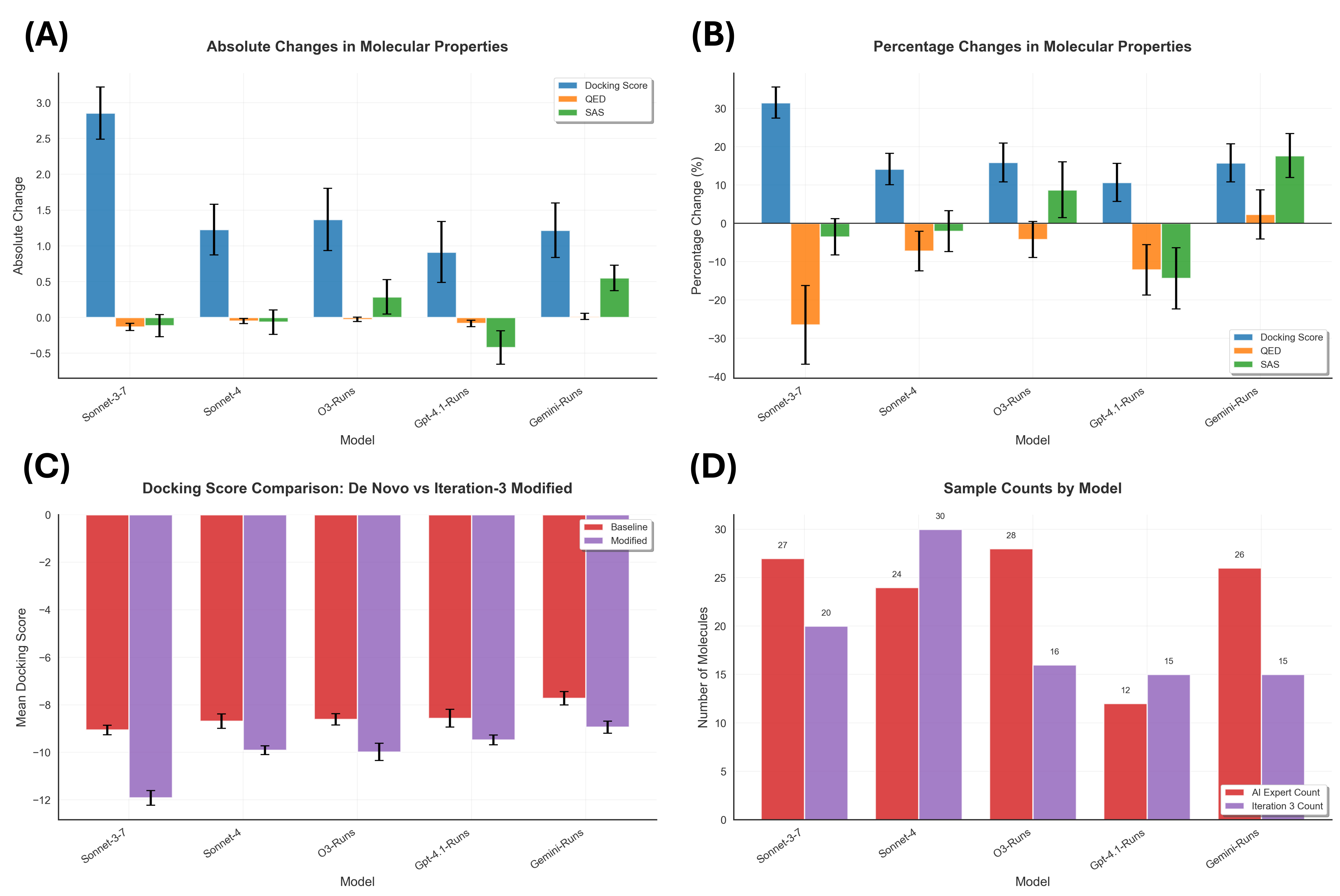}

    \caption{Absolute \textbf{(A)} and percent \textbf{(B)} changes in mean docking score, QED, and SAS between the AI-expert baseline molecules and the Iteration-3 analogues generated by each LLM-driven agent team. Error bars denote s.e.m. over three replicate runs. Panel \textbf{(C)} compares the raw docking means before and after optimisation, while panel \textbf{(D)} reports the number of molecules each model evaluated en route to its ten final proposals (no cap on intermediates was imposed). Lower docking scores and SAS, and higher QED, are desirable.
}
    \label{fig:img2}
\end{figure*}

Figure 3 summarizes the pair-wise Tanimoto similarities between the starting molecules (Database Agent and AI-Expert) and the analogues produced in the three Medicinal-Chemist (MC) rounds. Three practical points arise. \textbf{(i)} Degree of change across rounds. GPT-4.1 shows the smallest structural shift: its first MC round remains 0.76 similar to the AI-Expert scaffold and still 0.73 similar after the third round, indicating mainly local edits. Sonnet-4 behaves in a comparable, although slightly more permissive, manner (similarity drops to 0.54 by MC-3). In contrast, o3, Gemini, and Sonnet-3.7 move further away from their initial proposals, with AI-Expert MC-3 similarity falling to 0.30–0.36. \textbf{(ii)} Influence of the database compounds. For GPT-4.1, o3, Sonnet-3.7 and Sonnet-4, none of the MC analogues shares more than 0.30 similarity with any database molecule, suggesting limited reuse of pre-existing motifs. Gemini is the only model that temporarily returns toward the database set (0.49 similarity in MC-2), hinting that it occasionally revisits known chemotypes before continuing optimization. \textbf{(iii)} Exploration versus refinement. Ranking the models by the overall drop in similarity from AI-Expert to MC-3 gives the order o3 > Gemini , Sonnet-3.7 > Sonnet-4 > GPT-4.1. Thus, o3 explores the widest chemical space, while GPT-4.1 focuses on incremental refinement, with the remaining models occupying intermediate positions.

The heat-maps also reveal how the editing process settles over time. For every model the similarity between successive Medicinal-Chemist rounds is higher than the jump from the AI-Expert seed to MC-1, showing that the first round makes the largest structural leap and later rounds apply finer adjustments. This convergence is strongest for GPT-4.1 and Sonnet-4 (MC-2 vs. MC-3 = 0.56 and 0.79, respectively), while Gemini keeps introducing fresh changes later in the run (MC-2 vs. MC-3 = 0.52). Finally, Sonnet-3.7 stands out for starting from scaffolds that already resemble the small compound collection supplied by the Database Agent (DB vs. AI-Exp = 0.45), yet it still ends up almost as far from that starting space as Gemini and o3 after three optimisation rounds. Together with the points noted above, the figure shows how each LLM balances re-using familiar motifs against exploring new chemical space.

\begin{figure*}[h]

    \centering
    \includegraphics[width=\textwidth]{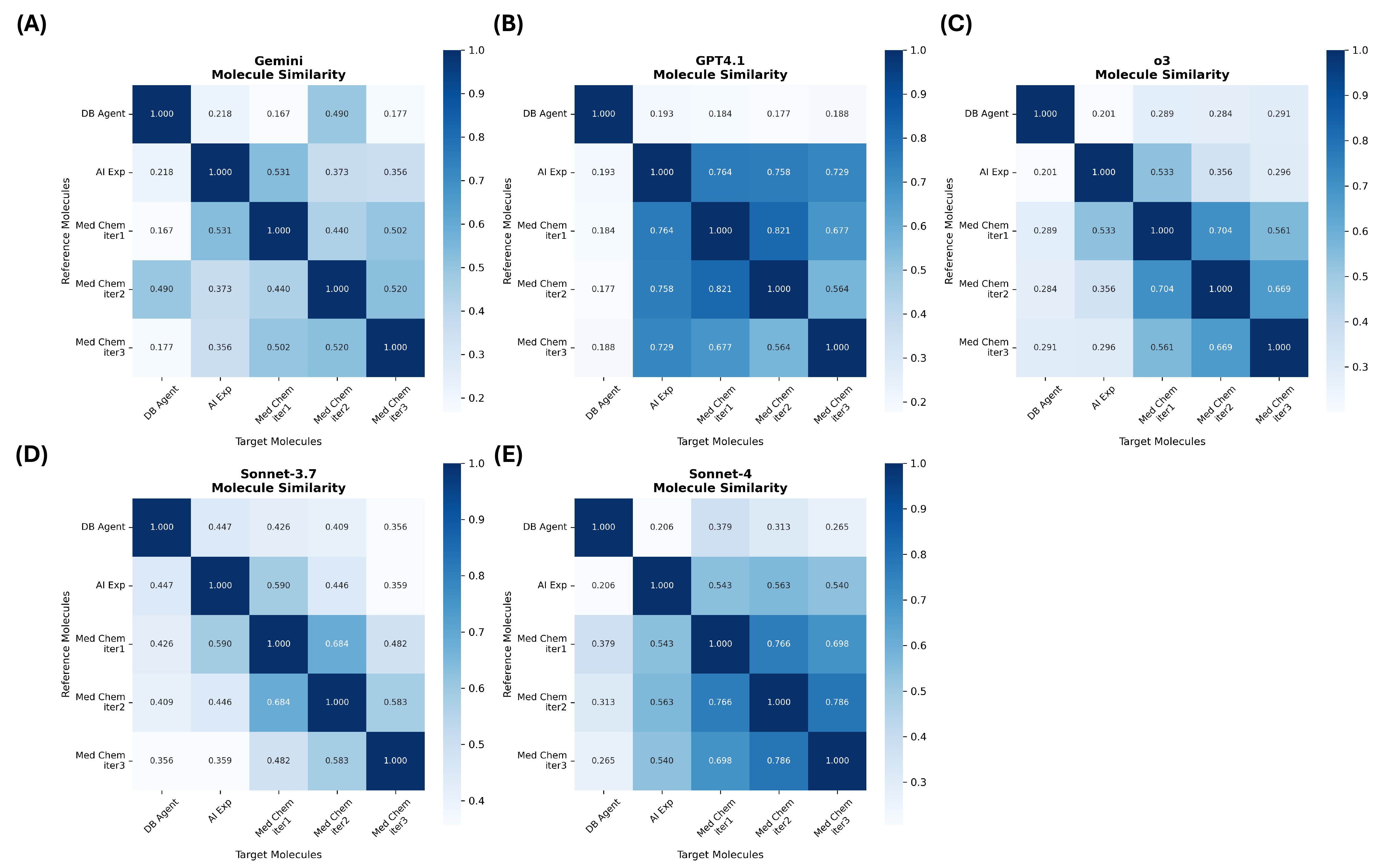}

    \caption{Heat-maps \textbf{(panels A–E)} report Tanimoto similarities among \textbf{(i)} the Database-Agent compounds, \textbf{(ii)} the AI-Expert de-novo proposals, and \textbf{(iii)} the molecules produced after three Medicinal-Chemist (MC) optimisation rounds for each LLM (Gemini, GPT-4.1, o3, Sonnet-3.7 and Sonnet-4). Darker squares indicate higher structural overlap (scale 0–1). Reading each map row-wise or column-wise shows \textbf{(1)} how much the first MC edit departs from the AI-Expert scaffold, \textbf{(2)} how far later MC rounds move away from earlier ones, and \textbf{(3)} the extent to which any of the models return to motifs present in the database compounds.
}
    \label{fig:img4}
\end{figure*}

We extended the comparison between the two Anthropic models from three replicates to twenty independent MAS runs to decrease the bias by increasing variance in Table 4. The larger sample size preserves the key pattern seen earlier while sharpening the trade-offs between the models. First, \textit{claude-sonnet-4-20250514} again explores far more chemical space 704 intermediates versus 509 for \textit{claude-3-7-sonnet-20250219}, yet this breadth comes at the cost of molecular validity: almost one-fifth of \textit{claude-sonnet-4-20250514}’s SMILES are invalid (validity is 0.803), whereas \textit{claude-3-7-sonnet-20250219} remains the high validity (0.998). \textit{claude-sonnet-4-20250514} generates with slightly higher internal diversity (0.844 ± 0.026 vs 0.791 ± 0.046). From a drug-likeness perspective \textit{claude-sonnet-4-20250514} performed better compared to other models. It delivers higher QED (0.590 ± 0.170), 97 \% Lipinski compliance, easier synthesis (SA = 3.01 ± 0.60), and a lower Fréchet ChemNet Distance (18.0 vs 20.1), indicating physicochemical profiles closer to known AKT1 inhibitors. 

Crucially, target-specific performance still tilts in favour of \textit{claude-3-7-sonnet-20250219}. Its mean docking score remains about 1 kcal · mol\textsuperscript{-1} better (-10.28 ± 1.80 vs -9.28 ± 1.37), and nearly 70 \% of its compounds beat the stringent -9 kcal · mol\textsuperscript{-1} threshold, compared with 56 \% for \textit{claude-sonnet-4-20250514}. In other words, \textit{claude-3-7-sonnet-20250219} continues to translate its more focused (and syntactically cleaner) exploration strategy into higher hit rates, despite generating fewer and, by some metrics, less “drug-like” molecules overall.

Taken together with the three-run analysis, the twenty-run experiment reinforces our earlier conclusion: \textit{claude-3-7-sonnet-20250219} is the superior choice when docking affinity is the principal optimisation goal, whereas \textit{claude-sonnet-4-20250514} provides a richer and more drug-like molecules but at the expense of potency and SMILES validity. Which model to deploy therefore depends on whether the project values maximal binding energy or broader, medicinal-chemistry-friendly space.

\renewcommand{\theadfont}{\fontsize{7}{11}}
\begin{table}[h]
    \setlength\tabcolsep{10pt}
    \caption{Performance comparison of MAS runs with \textit{claude-3-7-sonnet-20250219} and \textit{claude-sonnet-4-20250514} models, twenty parallel runs are done to create more molecules to decrease the bias.}
    \centering
    \fontsize{7}{14}\selectfont
    \makebox[\textwidth]{%
    \begin{tabular}{lcc}
    \hline
    & \textbf{MAS with Sonnet-3.7} & \textbf{MAS with Sonnet-4} \\
    \hline
    \textbf{Generation Quality} \\ \hline
    Molecules Attempted (N)\textsuperscript{*} & 509 & 704 \\
    Validity (\%)                              & \textbf{0.998} & 0.803 \\
    Uniqueness (\% of Valid)                   & 0.972 & \textbf{0.988} \\
    Novelty (\% vs.\ Actives)                  & 1.000 & 1.000 \\
    Internal Diversity                         & 0.791\,$\pm$\,0.046 & \textbf{0.844}\,$\pm$\,0.026 \\
    Similarity to known inhibitors             & \textbf{0.433}\,$\pm$\,0.173 & 0.354\,$\pm$\,0.160 \\
    FCD                                         & 20.071 & \textbf{18.044} \\
    \hline
    \textbf{Drug-Likeness \& Properties} \\ \hline
    QED (Mean\,$\pm$\,SD)                      & 0.421\,$\pm$\,0.171 & \textbf{0.590}\,$\pm$\,0.170 \\
    Lipinski Ro5 Pass (\%)                     & 71.40 & \textbf{96.80} \\
    SA Score (Mean\,$\pm$\,SD)                 & 3.196\,$\pm$\,0.521 & \textbf{3.011}\,$\pm$\,0.601 \\
    LogP (Mean\,$\pm$\,SD)                     & 3.676\,$\pm$\,1.601 & \textbf{3.264}\,$\pm$\,1.248 \\
    \hline
    \textbf{Target-Specific Performance} \\ \hline
    Docking Score (Mean\,$\pm$\,SD)            & $-10.276$\,$\pm$\,1.802 & $-9.275$\,$\pm$\,1.369 \\
    Hits (Dock $< -9.0$ kcal/mol) (\%)         & \textbf{69.49} & 55.65 \\
    \hline
    \end{tabular}}
{\raggedright * We did not enforce a fixed exploration depth; each LLM was free to generate and discard as many intermediate structures as it deemed necessary before converging on its ten final proposals. Therefore, “Molecules Attempted (N)” reports the total number of candidate SMILES evaluated across all optimization steps (and all three replicates) for each model, and the values naturally differ from one LLM to the next.
\par} 
\vspace{0.5em}
\hrule
    \label{tab:mas_sonnet_metrics}
\end{table}

\subsection{Comparison of baseline LLM, Single-Agent, and Multi-Agent Setup}

To analyse the influence of agentic architecture on molecular optimization, we compared the performance of the \textit{claude-3-7-sonnet-20250219} model in three configurations: as a baseline LLM relying solely on its internal knowledge, a Single Agent wielding all five tools, and a Multi-Agent System (MAS) where tool use was distributed. The results reveal distinct strategic trade-offs rooted in how each system interacts with computational tools. The baseline LLM, lacking real-time computational feedback, still demonstrated internal molecular understanding capability, achieving the highest hit rate (98.0\%) and a strong average docking score (-10.65 kcal/mol). The Single Agent architecture, by actively using all tools, produced molecules with superior drug-likeness (100\% Lipinski pass rate) and diversity. However, this broad focus appeared to dilute its efforts, resulting in the worst average docking score (-9.72 kcal/mol). The two agentic architectures produced markedly different outcomes. The MAS configuration, by distributing tasks, proved highly effective at single-objective optimization, achieving the lowest average docking score (-10.97 kcal/mol). Conversely, the Single Agent, which must process all tool signals simultaneously, produced molecules with a more balanced profile, excelling in drug-likeness metrics despite a higher average docking score.

In Table 5, these divergent outcomes highlight that the true value of agentic systems lies in their ability to operationalize iterative feedback from tools. Each molecular modification is a hypothesis, and the tools provide empirical validation, critical signals that confirm success or failure. This trial-and-error process builds in-context knowledge, allowing the system to learn what works. The Single Agent's diminished potency suggests that receiving simultaneous, competing signals from five different tools creates a bottleneck, forcing a conservative strategy that compromises the primary objective, a phenomenon we explore as a "reasoning bottleneck". The MAS circumvents this by creating focused feedback loops; its distributed design assigns specific tools to different agents, isolating their signals. This allows one agent to aggressively optimize for potency based on docking scores while other agents can focus on their goals. This division of labor drives highly focused target-specific optimization and creates a transparent audit trail of the optimization logic, making the system's strategy interpretable. Ultimately, the choice of architecture dictates the optimization strategy. The MAS proves most effective for aggressively pursuing a single objective like predicted binding affinity, while the Single-Agent architecture naturally adopts a more balanced approach, prioritizing overall drug-likeness.

\renewcommand{\theadfont}{\fontsize{7}{11}}
\begin{table}[h]
    \setlength\tabcolsep{10pt}
    \caption{Comparison of baseline LLM, single-agent, and multi-agent setup on generative metrics, drug-likeness, and target specific performance, using \textit{claude-3-7-sonnet-20250219} LLM as backing model. Runs repeated three times for each setup.}
    \centering
    \fontsize{7}{14}\selectfont
    \makebox[\linewidth]{%
    \begin{tabular}{lccc}
    \hline
    & \textbf{Baseline LLM} & \textbf{Single Agent} & \textbf{MAS} \\
    \hline
    \textbf{Generation Quality} \\ \hline
    Molecules Attempted (N)\textsuperscript{*} & 101 & 63 & 75 \\
    Validity (\%)                              & \textbf{1.000} & 0.984 & \textbf{1.000} \\
    Uniqueness (\% of Valid)                   & 0.970 & \textbf{1.000} & \textbf{1.000} \\
    Novelty (\% vs.\ Actives)                  & \textbf{1.000} & \textbf{1.000} & \textbf{1.000} \\
    Internal Diversity                         & 0.745\,$\pm$\,0.029 & \textbf{0.839}\,$\pm$\,0.028 & 0.764\,$\pm$\,0.056 \\
    Similarity to known inhibitors             & 0.315\,$\pm$\,0.041 & 0.260\,$\pm$\,0.045 & \textbf{0.458}\,$\pm$\,0.163 \\
    FCD                                         & 32.914 & 28.926 & \textbf{26.148} \\
    \hline
    \textbf{Drug-Likeness \& Properties} \\ \hline
    QED (Mean\,$\pm$\,SD)                      & 0.428\,$\pm$\,0.098 & \textbf{0.615}\,$\pm$\,0.139 & 0.395\,$\pm$\,0.177 \\
    Lipinski Ro5 Pass (\%)                     & 81.30 & \textbf{100} & 56.00 \\
    SA Score (Mean\,$\pm$\,SD)                 & \textbf{2.525}\,$\pm$\,0.267 & 3.062\,$\pm$\,0.538 & 3.264\,$\pm$\,0.403 \\
    LogP (Mean\,$\pm$\,SD)                     & 4.123\,$\pm$\,1.033 & \textbf{3.372}\,$\pm$\,0.865 & 3.667\,$\pm$\,1.599 \\
    \hline
    \textbf{Target-Specific Performance} \\ \hline
    Docking Score (Mean\,$\pm$\,SD)            & $-10.648$\,$\pm$\,0.855 & $-9.718$\,$\pm$\,0.991 & \textbf{$-10.973$}\,$\pm$\,1.573 \\
    Hits (Dock $< -9.0$ kcal/mol) (\%)         & \textbf{97.96} & 67.44 & 96.00 \\
    \hline
    \end{tabular}}
{\raggedright * We did not enforce a fixed exploration depth; each LLM was free to generate and discard as many intermediate structures as it deemed necessary before converging on its ten final proposals. Therefore, “Molecules Attempted (N)” reports the total number of candidate SMILES evaluated across all optimization steps (and all three replicates) for each model, and the values naturally differ from one LLM to the next.
\par} 
\vspace{0.5em}
\hrule
    \label{tab:baseline_single_mas_metrics}
\end{table}

\subsection{Iteration Based Improvement on Molecules}

Figure 5 presents the outcomes of the iterative optimization process by the Medicinal Chemist (MC) agent, initiating from molecules generated by the de novo molecules generation model of the AI Expert model. The agent's focus on its primary objective is immediately evident in the docking scores (A). With each iteration, the agent successfully drove the mean predicted binding affinity lower, from an initial -10.05 kcal/mol to a final -11.91 kcal/mol, a 31.5\% improvement over the baseline. This demonstrates the agent's profound capability to leverage the docking tool for targeted optimization. The best individual MC docking score also showed strong improvement, from -13.04 in iteration 1 to a remarkable -14.72 in iteration 2, before settling at -13.63 in iteration 3. However, this gain in docking performance was accompanied by a decrease in the mean Quantitative Estimate of Drug-likeness (QED, B). The MC's mean QED started at 0.404 (n=29), remained stable at 0.404 (n=26) in iteration 2, and then declined to 0.368 (n=20) in iteration 3. This trajectory resulted in a 26.5\% reduction in QED compared to the AI Expert's mean baseline (visually plotted at ~0.78). The mean Synthetic Accessibility Score (SAS, C) for MC-modified molecules fluctuated, starting at 3.26 (n=29), decreasing slightly to 3.20 (n=26) in iteration 2, and then increasing to 3.36 (n=20) in iteration 3. This final SAS value indicated a 3.5\% worsening in synthetic feasibility relative to the AI Expert's mean SAS (visually plotted at ~2.1). LogP values (D) exhibited considerable variation: the mean increased from 3.57 (n=29) at iteration 1 to a peak of 3.94 (n=26) at iteration 2, before decreasing to 3.44 (n=20) at iteration 3. The results for sonnet-3-7 highlight the MC agent's strong capability to optimize the primary target (docking score), though this can involve trade-offs, leading to reductions in QED and slight increases in SAS difficulty compared to the AI Expert's initial molecular profile.

\begin{figure*}[h]

    \centering
    \includegraphics[width=\textwidth]{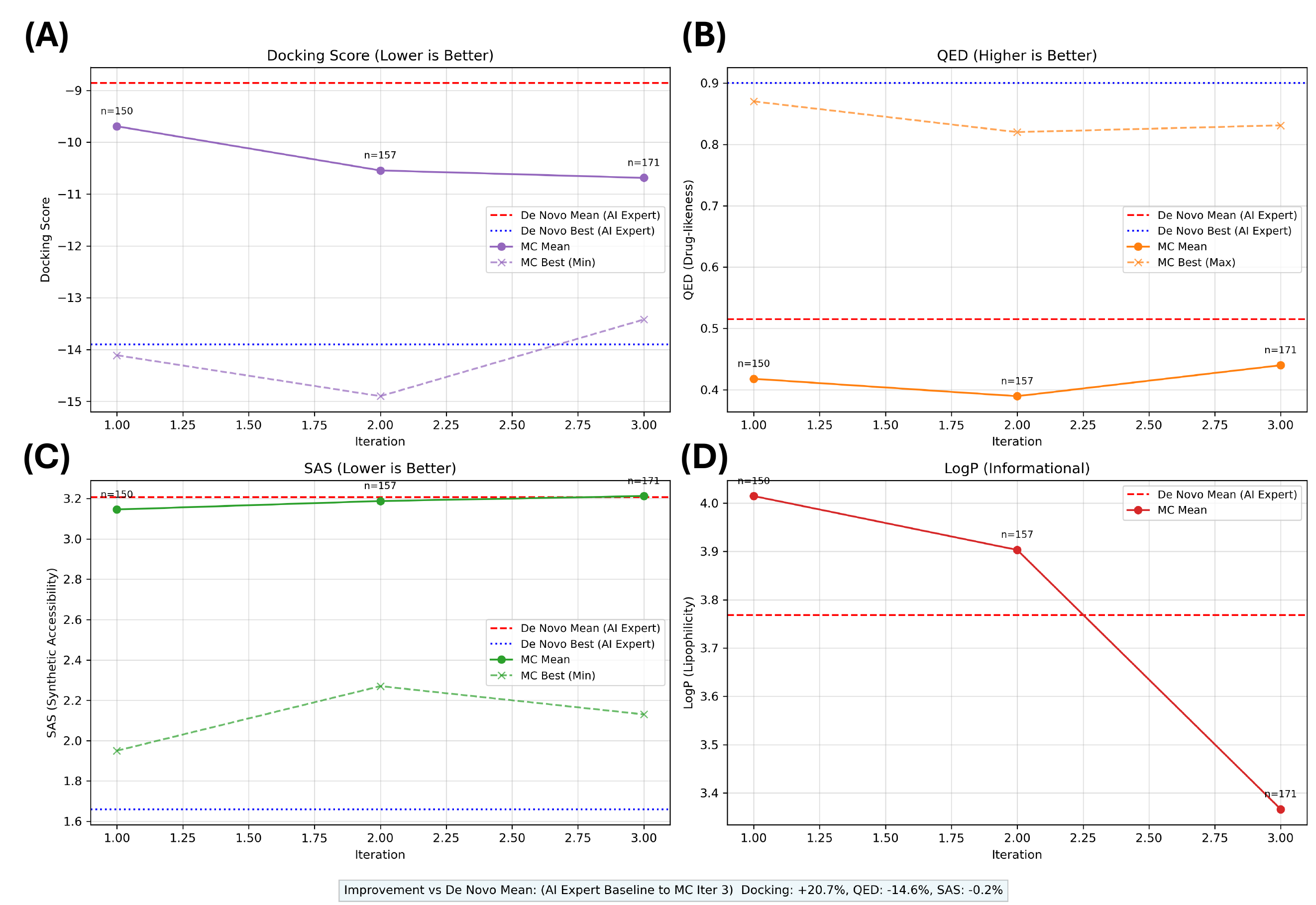}

    \caption{Iterative molecular property optimization by a Medicinal Chemist (MC) agent versus a de novo AI Expert baseline (sonnet-3-7). The subplots depict: \textbf{(A)} Docking Score (lower is better), \textbf{(B)} Quantitative Estimate of Drug-likeness (QED; higher is better), \textbf{(C)} Synthetic Accessibility Score (SAS; lower is better), and \textbf{(D)} LogP (lipophilicity; informational). Mean MC performance (solid lines) and best individual MC molecule performance (dashed lines) are tracked across three iterations and compared against the AI Expert's mean (blue dotted line) and best (red dotted line) initial outputs. The number of molecules (n) per stage is noted. The overall percentage change from the AI Expert baseline mean to the MC's iteration 3 mean is summarized in the footer.
}
    \label{fig:img2}
\end{figure*}
\subsection{Molecular Modifications}

This modification sequence provides a powerful case study in the agent's capacity to iteratively modify structures based on computational feedback. The process began with the agent identifying a potential liability in the de novo molecule: a complex triazole-containing system that could pose a synthetic challenge. In one modification step, the system generated a molecule where this moiety was simplified to an N-methylpyrazole where it improved drug-like properties. The outcome was a deliberate and informative trade-off: as predicted, the synthetic accessibility and QED improved significantly (SAS: 2.670, QED: 0.773), but this came at the direct cost of the primary objective, with predicted binding affinity worsening to -9.32 kcal·mol\textsuperscript{-1}. Modifications steps are represented in Figure 4.

\vspace{1em}
Crucially, the agent did not discard this failed experiment. Instead, it retained the previously successful modification to the pyrazole ring while introducing a new change elsewhere, demonstrating a multi-step modification pathway. It identified the N-methylpyrazole as a successful modification for improving SAS and retained it. To solve the new problem of poor potency, it executed a distinct and complementary strategy: introducing a morpholine group via a linker to explore new binding interactions. This composite strategy was a resounding success. The final molecule not only recovered the lost potency but surpassed the original, achieving a better docking score of -11.09 kcal·mol\textsuperscript{-1} while preserving the synthetic accessibility gained in the first step. This ability to learn from a suboptimal result, isolate the successful components of a change, and layer on a new, targeted solution showcases a multi-step reasoning process.

\vspace{1em}

The screening began with a de novo scaffold that bound AKT1 at -9.73 kcal mol\textsuperscript{-1} while preserving good drug-likeness (QED 0.618). A first optimization path introduced an oxadiazolone–quinazoline core and exchanged the piperazine for piperidine (modifications (1)–(3)). This cut synthetic complexity and deepened binding to -10.68 kcal mol\textsuperscript{-1}, although QED fell to 0.481. A second, parallel path swapped the thiophene for a hydroxyphenyl ring and N-methylated the piperazine (modifications (4)–(5)). Binding was retained (-10.0 kcal mol\textsuperscript{-1}) and drug-likeness improved markedly (QED 0.84), demonstrating that the hydroxyphenyl motif can replace the heteroaromatic ring without loss of potency. \\

In the next cycle, the oxadiazolone series was fluorinated. Addition of difluoromethyl and difluoroethyl groups (modifications (6) and (8)) lifted affinity to -10.71 and -11.12 kcal mol\textsuperscript{-1}, respectively, by reinforcing hydrophobic contacts and maintaining the ASP292 salt bridge. The cost was a lower QED (0.300–0.442), reflecting increased size and lipophilicity. A related analogue bearing both fluorinated side chains (modification (9)) kept the strong binding of the parent and suggested enhanced metabolic stability. For the hydroxyphenyl branch, trimming a methyl group (modification (10)) yielded the most drug-like profile in the set (QED 0.863) with only a small loss in docking score (-9.33 kcal mol\textsuperscript{-1}). \\

Taken together, these examples demonstrate that the LLM is not making random changes; the proposed modifications are non-random and consistent with common medicinal chemistry heuristics, suggesting the LLM's outputs are guided by patterns in its training data. These modification sequences, tracked through the system's provenance records, provide a clear audit trail of the agent's reasoning. These results demonstrate that adding tool-based signals at test time strengthens an LLM’s in-context learning, enabling it to infer facts that lie beyond its built-in molecular knowledge. \\

\begin{figure*}[h]

    \centering
    \includegraphics[width=\textwidth]{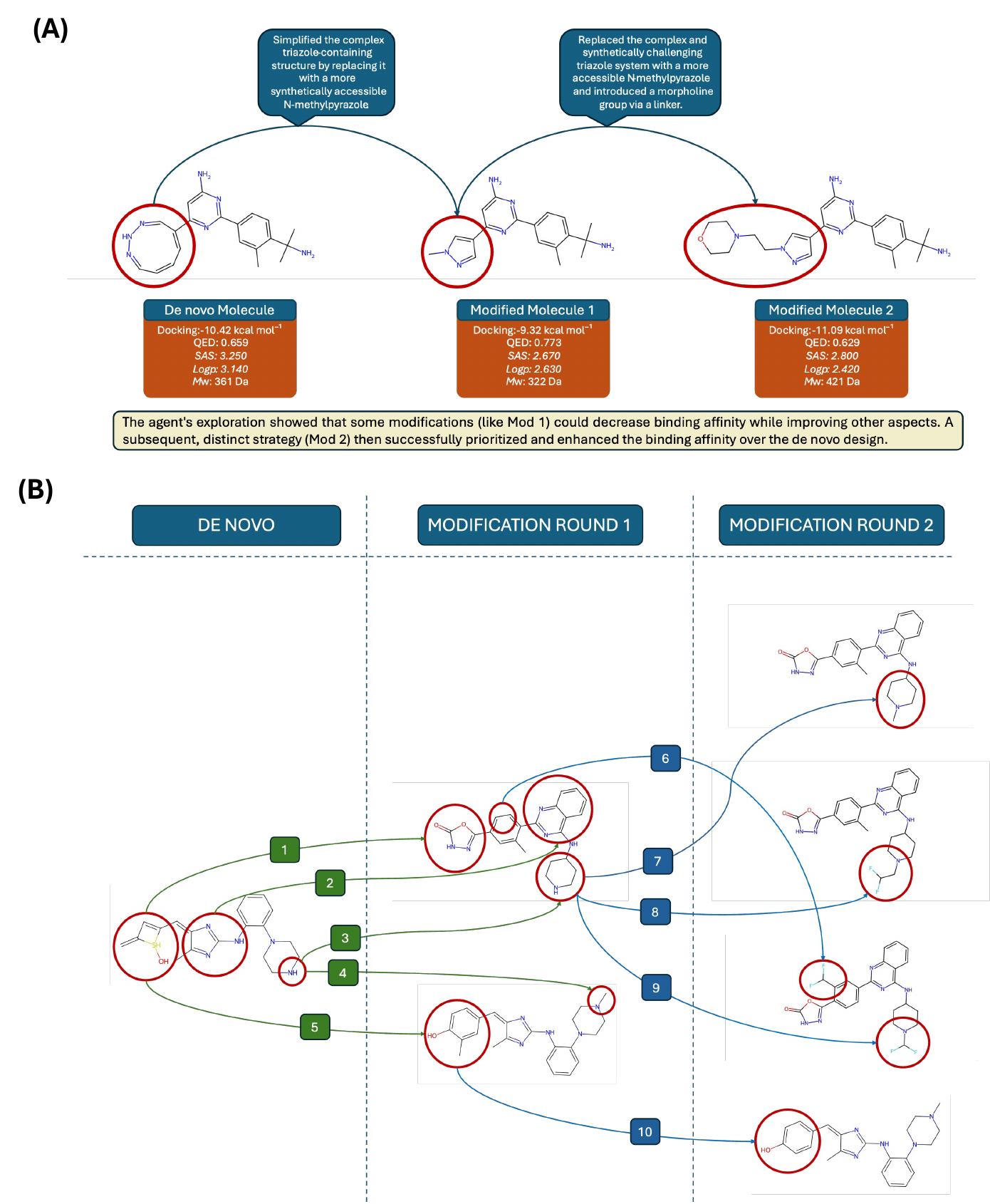}
    \label{fig:img3}
\end{figure*}
\clearpage
\captionof{figure}{\textbf{(A)} LLM-based agent's iterative molecular modification strategy, showcasing its ability to learn from outcomes and adapt its approach. The process starts with a "De novo Molecule" (left) featuring a complex triazole-containing system. The first attempt, "Modified Molecule 1" (center), simplifies this to an N-methylpyrazole, improving Quantitative Estimate of Drug-likeness (QED: 0.659 to 0.773) and Synthetic Accessibility Score (SAS: 3.250 to 2.670) but at the cost of predicted binding affinity (Docking: -10.42 to -9.32 kcal/mol). Learning from this, "Modified Molecule 2" (right) retains the beneficial N-methylpyrazole substitution while introducing a morpholine group via a linker. This second distinct strategy successfully enhances the docking score to -11.09 kcal/mol, surpassing the original de novo molecule, while maintaining an improved SAS (2.800) and acceptable QED (0.629). This demonstrates the agent's capacity to prioritize different objectives and build upon successful modifications. \textbf{(B)} The LLM agent's branching optimization strategy showcases molecular reasoning capacity. The process initiates from a single de novo scaffold (left), which the agent diversifies in "MODIFICATION ROUND 1" by proposing multiple distinct analogs. Each new molecule targets different regions of the parent structure (highlighted in red) to address specific liabilities or explore new binding interactions. In "MODIFICATION ROUND 2," the agent demonstrates iterative learning by selecting promising intermediates from Round 1 and refining them further. This branching workflow visualizes the agent's ability to run parallel optimization trajectories, balancing competing objectives like potency, synthesizability, and drug-likeness showcasing in-context learning capacity of LLMs.} 

\section{Discussion}

This study systematically investigated the impact of agentic architecture on the performance of LLMs in a tool-augmented, target-based molecular optimization task. Our results provide a clear narrative: the architectural configuration of an agent-based system fundamentally dictates its strategic biases and ultimate success. The key finding is the distinct trade-off between optimizing for a primary objective, such as binding potency, and maintaining broad, favorable drug-like properties. While all configurations demonstrated the capacity for molecular design, they achieved it through markedly different strategies, revealing crucial insights into how to effectively structure AI for complex scientific problems. \\

The most striking result was the divergent strategic behavior of the Single Agent and Multi Agent System (MAS). The MAS demonstrated a clear capacity for aggressive single-vector optimization, achieving the lowest average docking scores. In parallel, the Single Agent's strategy resulted in molecules with superior drug-likeness, highlighting a fundamental divergence in how architectural design prioritizes competing scientific objectives. In contrast, the Single Agent, while less effective at maximizing potency, excelled in producing molecules with better drug-like characteristics, evidenced by its high Lipinski compliance and the highest average QED score. This suggests that the MAS’s success stems from its division of labor. By assigning specialized tools to distinct agents, the system decomposes the complex multi-objective problem. The Medicinal Chemist agent, for instance, could operate within a focused feedback loop, using signals from its docking tool to aggressively pursue potency without being constrained by conflicting signals from other property calculators using the advantage of in-context learning through reasoning trajectories. The baseline LLM also demonstrated formidable performance, achieving a high hit rate and potent docking scores, highlighting the powerful intrinsic chemical knowledge embedded within modern foundation models. However, its exploration lacked the guided, iterative refinement and interpretability provided by the structured agentic frameworks. \\

Molecular similarity heatmaps in Figure 3 reveals two distinct exploration regimes across the tested LLMs. GPT-4.1 and Sonnet-4 retain > 0.70 Tanimoto similarity to their AI-Expert seeds even after three optimisation rounds, echoing evidence that standard autoregressive models “think too fast,” privileging early uncertainty cues and therefore favouring conservative, low-risk edits \cite{pan2025large}. In contrast, o3 and Gemini show markedly larger scaffold shifts. A parallel pattern is reported in bandit benchmarks, where supplying the model with structured summaries of past interactions or explicit exploitation–exploration statistics at inference time leads to significantly stronger exploratory behaviour \cite{nie2025evolve}. Taken together, the heat-maps function as a proxy for each model’s position on the explore–exploit spectrum. A similarity drop signals an internal policy that values exploratory gain, whereas a shallow drop indicates a bias toward rapid but local refinement. However, why LLMs exhibit such divergent patterns under an identical agent environment remains unclear and requires further investigation.

We propose that the Single Agent's compromised potency can be attributed to a "reasoning bottleneck". When a single entity is tasked with concurrently managing and interpreting signals from five distinct tools, it faces competing and often contradictory imperatives, for example, the need to increase molecular complexity to improve binding versus the need to simplify it to improve synthetic accessibility. We hypothesize that when a single agent faces multi-parameter complexity, it naturally adopts a conservative, balanced strategy. The MAS architecture, by contrast, circumvents this need for balancing by isolating feedback loops. This is not necessarily overcoming a 'bottleneck,' but rather enabling a different, more focused optimization strategy that prioritizes excellence in a single metric, albeit at the potential expense of others like drug-likeness. Its hierarchical and specialized structure isolates these feedback loops, allowing for a more focused and aggressive optimization strategy toward the primary goal defined by the Principal Researcher. \\

Beyond performance metrics, a critical advantage of the agentic frameworks is their inherent interpretability. Unlike a monolithic, black-box model, the explicit communication between agents and their tool calls creates a transparent audit trail of reasoning trajectories, which allows us to analyze the system's reasoning process. Our case study of an iterative modification (Figure 3) provides a powerful example of this. We observed the agent executing a multi-step strategy: it first identified a liability, proposed a modification that successfully improved synthesizability but worsened potency, and then, learning from this suboptimal outcome, retained the beneficial change while layering on a new, distinct modification to recover and ultimately surpass the original molecule's predicted binding affinity. This ability to decouple problems and learn from intermediate failures showcases a complex, multi-step process that is crucial for effective multi-agent collaboration. \\

Situating our work in the broader context, studies like ChemCrow \cite{m2024augmenting} have demonstrated that LLMs can use chemistry tools. Our direct, controlled comparison of baseline, single-agent, and multi-agent architectures provides a granular analysis of these architectural trade-offs. While more ambitious end-to-end systems like PharmAgents \cite{gao2025pharmagents} aim to automate the entire discovery pipeline, our deliberate focus on the molecular optimization stage allows for a deeper and more controlled investigation of this critical phase, offering valuable insights for designing next-generation scientific automation platforms. \\

Nevertheless, this study has several limitations that frame avenues for future work. First, our toolset was primarily focused on predicted binding affinity and basic physicochemical properties. The absence of tools for predicting ADMET properties, selectivity, or metabolic stability likely contributed to the MAS’s tendency to produce potent but less drug-like molecules. A more comprehensive tool library would be necessary to navigate the true multi-objective landscape of drug design. Second, our findings are based on a single protein target, AKT1, and may not generalize to all target classes. Third, we acknowledge that docking score is an imperfect proxy \cite{paggi2024art} for true predicted binding affinity and biological activity. Finally, our system employed a fixed, sequential workflow, whereas real-world research is often more dynamic and parallelized. \\

These limitations point toward several exciting future directions. The development of more flexible, dynamic agentic systems, where agent teams can be assembled on-the-fly or collaborate in parallel based on the specific problem, presents a promising research frontier. Integrating more sophisticated computational tools to create richer, more realistic feedback loops will be essential for tackling later-stage optimization challenges \cite{swanson2024virtual}. Next-generation structure predictors such as AlphaFold-3 \cite{abramson2024accurate}, which natively co-fold proteins with small molecules, and Boltz-2 \cite{passaro2025boltz} which can predict predicted binding affinity of protein and ligand, could be exposed to the agents as additional tools. These signals would complement docking by providing higher-resolution poses and more reliable affinity estimates, reducing the risk of over-optimising against a single scoring function. Furthermore, the inherent interpretability of our MAS framework makes it an ideal candidate for human-in-the-loop workflows, where a human expert can intervene to guide the strategy, resolve conflicts, or validate novel hypotheses generated by the agents. In conclusion, this work demonstrates that the careful design of agentic architectures is a critical consideration for effectively leveraging LLMs in scientific workflows. By decomposing complex problems and assigning specialized tools to a team of agents, we can create powerful, effective, and interpretable systems capable of accelerating the computational design and optimization phase of scientific discovery. Lastly, by swapping in sequence-based generators and property predictors, the agents could design peptides, PROTAC linkers, RNA binders, or other biomolecular modalities, opening a path toward a unified, multi-modal optimisation platform. \\

\section{Code availability}

The source code and ready-to-use pipelines are available in the archived repository and can be accessed at https://github.com/deltawave-tech/delta. \\

\section{Acknowledgments}

We thank Tunca Doğan for his insightful comments and valuable discussions on the project.

\section{Author Information \& Contributions}

AU: Atabey Ünlü (atabey@deltawave.fr), \\

PR: Phil Rohr (phil@deltawave.fr), \\

AC: Ahmet Celebi (ahmet@deltawave.fr), \\

AU, PR, and AC conceptualized the study and designed the research-cycle pipeline. PR handled tool deployment and built the molecular-provenance system, while AU and AC implemented the pipeline and optimized its code. AU conducted the experiments and performed the downstream analyses. All three authors AU, PR, and AC, collaboratively wrote the manuscript.

%Bibliography
\bibliographystyle{unsrt}  
\bibliography{main}

\end{document}

\maketitle

\vspace{-8em}
\section*{Supplementary Text}

\subsection*{S1. System Architecture and Implementation Details}

\subsubsection*{S1.1. Framework and Software}

Our multi-agent system is implemented in Python. To ensure modularity and interoperability with various LLMs, we utilized LiteLLM. This library acts as a standardized interface, allowing us to interact with over different LLM APIs from providers like Anthropic \cite{anthropic}, OpenAI \cite{openai}, and Google \cite{gemini} using a consistent format. This approach decouples our agentic logic from any single model provider, facilitating direct model comparisons and future-proofing the system. We chose to build our framework directly on LiteLLM rather than using higher-level agent libraries like LangChain, as this provided the granular control necessary to implement our specific context management strategies and workflow logic.

\subsubsection*{S1.2. Parallel Execution and Scalability}

To efficiently explore the vast chemical space and ensure the robustness of our findings, the system employs a Parallel Execution and Best-of-N Selection strategy. Instead of running a single, long discovery campaign, we execute multiple (e.g., N=20) independent, full-cycle discovery pipelines simultaneously. Each parallel run operates as a self-contained experiment. Upon completion, the final candidate molecules from all N runs are aggregated. A final selection process then identifies the most promising compounds based on the primary optimization objective (e.g., docking score). This parallelized approach allows the system to explore diverse regions of chemical space concurrently, significantly accelerating the discovery of high-quality candidates compared to a purely linear search.

\subsection*{S2. Context Management for Efficient Agent Communication}

Standard agentic systems are often limited by LLM context window sizes. Passing the entire conversation history, including verbose tool outputs, is computationally expensive and can dilute the agent's focus on the primary objective. Our system implements a context management strategy to overcome this challenge.

\paragraph{Agent Communication}We implemented a synchronous, turn-based multi-agent architecture for in-silico drug discovery in which every agent is a large-language model augmented with a small, predefined tool set. All agents read from and write to a single chronological transcript that contains natural-language messages, raw tool outputs and a machine-readable projection produced on the fly by a dedicated Summary Parser. Each research cycle begins when the Principal Researcher inserts a strategic prompt that defines the iteration goals and constraints. Domain specialists, Database, Literature, AI Expert, Medicinal Chemist, Ranking and Scientific Critic, then act in a fixed sequence, using their tools to retrieve protein data, generate molecules, evaluate docking results or expose logical flaws. After each contribution the Summary Parser converts essential facts into a provenance system, preserving reasoning trajectories, molecule transformations, metadata like protein sequences, PDB paths and molecule identifiers across rounds. Once all specialists have spoken, the Principal Researcher synthesises the free-text discussion with the structured record, issues a concise summary and formulates objectives for the next iteration, thereby preventing unbounded context growth. The loop is repeated until a final report containing the ten highest-ranked candidate molecules and supporting evidence is produced, with all intermediate steps logged for reproducibility and auditability, ensuring transparent, data-driven decision making.

\paragraph{Tool Output Isolation and Forced Summarization.} The raw, verbose output from a tool (e.g., a full docking log file) is made available only to the agent that called the tool and only for its current turn. This detailed output is never appended to the global, shared conversation history. This design choice forces the agent to perform a cognitive step: it must process and interpret the raw data to extract the most critical findings. The agent is then prompted to formulate a concise, accurate summary of these findings, which is the only part added to the shared history. This transforms the shared history from a raw data dump into a curated log of scientific insights, preserving context window space and maintaining focus.

\paragraph{Iterative History Summarization.} To further enhance efficiency, the system does not pass the full, multi-turn history from previous cycles into a new one. Instead, at the start of a new research cycle, the agents are provided only with the final summary and overarching objective written by the Principal Researcher from the preceding cycle. This method proved to be highly performant and cost-effective, allowing the system to build upon previous conclusions without being encumbered by excessive detail.

\subsection*{S3. Molecular Provenance and Lineage Tracking}

To ensure transparent and reproducible outcomes, we engineered an in-memory provenance service that logs the evolutionary history of all molecular candidates. This service models the relationships between molecules and the transformations that generate them.

The system's core is a directed hypergraph, where transformations (agent-driven modifications) act as hyperedges. These hyperedges connect a set of input molecules to a set of output molecules. This structure precisely captures complex events common in molecular design, such as a single parent molecule yielding multiple derivatives in a branching optimization strategy.

Superimposed on this dependency graph are two distinct ordering structures: \textbf{(i)} A totally ordered temporal chain records the immutable, linear sequence of all transformations as they occur within a session. This provides a complete chronological account of every action taken by every agent. \textbf{(ii)} A direct lineage forest is maintained across the set of all molecules. In this structure, each molecule explicitly references its single parent, enabling the rapid traversal of its direct ancestry.

Each transformation in the temporal chain is attributed to the specific agent that originated it. This tripartite structure, a dependency hypergraph for data-flow provides a robust and auditable record of the entire design process. This forms the foundation for analyzing agent behavior, justifying design choices, and ensuring the logical integrity of the results.

\subsection*{S4. Agent Definitions and Prompts}

When creating the definitions of the agents, we followed the same style as Virtual Lab \cite{VirtualLab} where we prompted it with title, expertise, goal, and role. All the definitions and prompts of the agents are shared in the github repository. 

\subsection*{S5. Experiment Definitions}

\subsubsection*{S5.1. Comparative Evaluation of Large Language Models}

To identify the most suitable LLM for our molecular optimization tasks, we conducted an initial comparative study. We evaluated five distinct closed-source LLMs from three providers: OpenAI (gpt-4.1-turbo, o3) \cite{openai}, Anthropic (Claude 3 Sonnet, hereafter "Sonnet-3.7", and Claude 4 Sonnet as “Sonnet-4) \cite{anthropic}, and Google (Gemini 2.5 Pro, hereafter "Gemini") \cite{gemini}. Each of the five LLMs was integrated into the full Multi-Agent System (MAS). The system was tasked with generating and optimizing molecules targeting AKT1. Each model was run for three independent replicates. At the conclusion of each run, the model was prompted to propose ten final optimized molecules.The analysis included all unique molecules generated throughout the optimization process. The top-performing LLMs were selected for subsequent, more detailed experiments based primarily on their ability to achieve low docking scores and high hit rates.

\subsubsection*{S5.2. Scaled Performance Analysis of Top-Performing LLMs}

Based on the initial screening, the two top-performing models, Sonnet-3.7 and Sonnet-4, were selected for a more robust head-to-head comparison. The experimental setup was identical to the initial evaluation, with each model operating within the MAS to optimize molecules for AKT1. To increase statistical power and account for variability in model outputs, the number of independent replicates was increased from three to twenty runs for each model to decrease bias and increase the variance.

\subsubsection*{S5.3. Evaluation of Agentic Architectures}

To dissect the impact of system design on performance, we compared three distinct architectural configurations using the best-performing LLM, Sonnet-3.7. Each configuration was tested in three independent replicates. The LLM operated without access to any of the five external computational tools. It was tasked with generating and optimizing molecules based solely on its internal, pre-trained knowledge and the initial prompt. The LLM acted as a single agentic controller with direct access to all five computational tools. It could call any tool at any time to inform its molecular modification decisions. The LLM was integrated into our standard pipeline where responsibility for tool use was distributed across specialized agents. This modular design created focused feedback loops, allowing different agents to prioritize different signals (e.g., one agent focusing on docking scores, another on synthesizability).

\subsubsection*{S5.4. Analysis of Iterative Optimization by the Medicinal Chemist Agent}

To characterize the step-by-step optimization capability of the system, a detailed analysis of the iterative refinement process was conducted. This experiment was designed to isolate and evaluate the performance of the Medicinal Chemist (MC) agent, which is primarily responsible for improving binding affinity. The analysis focused on the MC agent powered by the Sonnet-3.7 LLM, which was identified as the top-performing model for docking score optimization in our initial comparisons. The process began with a set of de novo molecules generated by the AI Expert agent from a separate run. These molecules served as the starting point, or "Iteration 0" baseline, for the MC agent's refinement task. The MC agent was tasked with iteratively modifying this initial molecular set over three successive optimization rounds. In each round, the agent received the molecules from the previous stage, proposed modifications with the explicit goal of improving the docking score, and utilized the docking tool to evaluate the results. The number of molecules (n) carried forward to the next iteration was determined by the agent's internal selection logic. The performance trajectory of the MC agent across the three iterations was benchmarked against the mean and best property values of the initial baseline molecule set from the AI Expert agent. The overall optimization efficiency was quantified by calculating the final percentage change between the mean baseline values and the mean values observed after the third and final iteration. The results of this analysis are visualized in Figure 2.

\subsubsection*{S5.5. Analysis of Optimization Trajectories and Molecular Modifications}

To understand the step-by-step optimization process, we performed a detailed analysis of the MAS runs driven by the Sonnet-3.7 model. We tracked the evolution of key molecular properties across three sequential optimization iterations performed by the "Medicinal Chemist" agent. The mean and best values for docking score, QED, SA score, and LogP were plotted for each iteration and compared against the properties of the initial de novo molecules. We conducted a case-study analysis of specific molecular modification sequences. By visualizing the changes between the de novo starting material and the agent-modified analogs (Figures 3 \& 4), we evaluated the agent's chemical reasoning. This included assessing its ability to identify and address molecular liabilities, balance competing objectives (e.g., potency vs. synthesizability), and explore multiple optimization pathways in parallel from a single scaffold.

\subsection*{S6. Evaluation Metrics}

To comprehensively assess the performance of the LLMs and agentic architectures, a suite of metrics was employed, covering generation quality, chemical space exploration, drug-likeness, and target-specific efficacy. All molecular property calculations were performed using the RDKit cheminformatics library.

\paragraph{Validity.} This metric assesses the syntactic and chemical correctness of the generated Simplified Molecular Input Line Entry System (SMILES) strings. A molecule was considered valid if its SMILES string could be successfully parsed and sanitized by RDKit.

\paragraph{Uniqueness.} This metric calculates the proportion of structurally distinct molecules within the set of all validly generated molecules for a given run. It was determined by converting all valid SMILES to their canonical representation and counting the number of unique structures. A high uniqueness value indicates the model avoids redundant generation.

\paragraph{Novelty} This metric measures the originality of the generated molecules by comparing them against a curated dataset of known active inhibitors for the target protein, AKT1. A molecule was classified as novel if its canonical SMILES representation was not present in this reference database.

\paragraph{Internal Diversity} This metric quantifies the structural variety within a set of generated molecules. It was calculated as the average pairwise Tanimoto dissimilarity (1 - Tanimoto similarity) based on Morgan fingerprints (radius 2). A higher value (closer to 1.0) indicates greater structural diversity, suggesting the model explored a broader range of chemical scaffolds.

\paragraph{Similarity to Known Inhibitors} To assess how closely the generated molecules resemble established AKT1 inhibitors, we calculated the Tanimoto similarity using Morgan fingerprints. 

\paragraph{Fréchet ChemNet Distance (FCD)} FCD measures the similarity between the distribution of generated molecules and a reference set (known AKT1 inhibitors) in terms of their chemical and physical properties. It leverages features from a pre-trained neural network to provide a more holistic comparison than structural similarity alone. A lower FCD score is better, indicating a higher similarity between the two sets.

\paragraph{Quantitative Estimate of Drug-likeness (QED)} QED is a metric that quantifies the drug-likeness of a molecule based on a desirability function of key physicochemical properties. It yields a score between 0 and 1, where higher values are better, indicating a greater resemblance to known oral medications \cite{QED}.

\paragraph{Lipinski's Rule of Five (Ro5)} This metric assesses the potential for oral bioavailability based on a set of well-established physicochemical guidelines. We report the percentage of molecules that pass the rule (i.e., have zero or one violation). A high pass rate is desirable \cite{lipinski}. 

\paragraph{Synthetic Accessibility (SA) Score} The SA score estimates the ease of synthesis for a given molecule on a scale from 1 (easy) to 10 (very difficult). The score is calculated based on fragment contributions and penalties for structural complexity. A lower SA score is better, indicating that a molecule is more likely to be synthetically feasible \cite{SA}.

\paragraph{LogP} This metric represents the logarithm of the octanol-water partition coefficient, a key measure of a molecule's lipophilicity. We compared generated values against the average LogP of known AKT1 inhibitors (around 4.0) as a benchmark \cite{logp}.

\paragraph{Docking Score} This metric provides a computational estimate of the binding affinity between a generated molecule and the active site of the target protein, AKT1. Scores are reported in kcal/mol, with more negative (i.e., lower) values indicating a stronger predicted binding interaction. This was the primary metric for evaluating optimization success.

\paragraph{Hits} This metric represents the percentage of generated molecules that achieved a docking score below a predefined threshold of -9.0 kcal/mol. It serves as a practical measure of success, quantifying the system's efficiency at generating potent binders.

\subsection*{S7. Reasoning Examples from History}

Each experiment run and full history will be shared in the github repository of the paper.